\documentclass[journal]{IEEEtran}

\usepackage[
  backend=biber,
  style=ieee
]{biblatex}
\AtBeginBibliography{\scriptsize}

\usepackage{amsmath,amssymb,amsfonts}
\usepackage{siunitx}
\usepackage{graphicx}
\usepackage[dvipsnames]{xcolor}
\usepackage[
  unicode,
  linkcolor={red!50!black},
  citecolor={blue!50!black},
  urlcolor={blue!80!black},
  colorlinks,
  hidelinks,
  bookmarks=false,
]{hyperref}
\usepackage[capitalise]{cleveref}
\usepackage{url}
\usepackage[shortcuts,acronym]{glossaries}
\glsdisablehyper
\usepackage[caption=false,font=footnotesize]{subfig}
\usepackage{tikz}
\usetikzlibrary{angles,quotes}
\usepackage{balance}

\newcommand{\DOI}{XX.XXXX/XXX.XXXX.XXXXXX} 
\usepackage[placement=top,vshift=-7]{background}
\SetBgScale{1.0}
\SetBgContents{Preprint version. Submitted to the IEEE for possible publication.}
\SetBgColor{black}
\SetBgAngle{0}
\SetBgOpacity{1.0}

\newcommand{\directivity}{\ensuremath{G}}
\newcommand{\directivityjoint}{\ensuremath{\mathcal{G}}}
\newcommand{\azimuth}{\ensuremath{\alpha}}
\newcommand{\inclination}{\ensuremath{\beta}}
\newcommand{\power}{\ensuremath{P}}
\newcommand{\wavelength}{\ensuremath{\lambda}}
\newcommand{\radiodistance}{\ensuremath{d}}

\newcommand{\state}{\ensuremath{\mathcal{S}}}
\newcommand{\statecontrol}{\ensuremath{\hat{\mathcal{S}}}}
\newcommand{\legendre}{\ensuremath{\mathcal{P}}}
\newcommand{\timephys}{\ensuremath{t}}
\newcommand{\roll}{\ensuremath{\omega^x}}
\newcommand{\pitch}{\ensuremath{\omega^y}}
\newcommand{\yaw}{\ensuremath{\omega^z}}

\newcommand{\features}{\ensuremath{\mathcal{F}}}

\newcommand{\numReal}{\ensuremath{\mathbb{R}}}

\renewcommand{\vec}[1]{{\boldsymbol{#1}}}
\newcommand{\fact}[1]{#1!}
\newcommand{\abs}[1]{{\vert#1\vert}}

\newcommand{\ivalcc}[1]{\ensuremath{\left<#1\right>}}
\newcommand{\transposesym}{\intercal}
\newcommand{\transpose}[1]{{#1}^\transposesym}

\DeclareMathOperator*{\argmin}{arg\,min}

\newcommand{\rx}{{rx}}
\newcommand{\tx}{{tx}}
\newcommand{\uava}{\ensuremath{a}}
\newcommand{\uavb}{\ensuremath{b}}
\newcommand{\uavc}{\ensuremath{c}}
\newcommand{\uavn}{\ensuremath{N}}
\newcommand{\parama}{\ensuremath{\param^a}}
\newcommand{\paramb}{\ensuremath{\param^b}}
\newcommand{\paramc}{\ensuremath{\param^c}}
\newcommand{\param}{\ensuremath{p}}
\newcommand{\uav}{\ensuremath{u}}

\newacronym{wrt}{w.r.t.}{with respect to}
\newacronym{dof}{DoF}{Degrees of Freedom}
\newacronym{uav}{UAV}{Uncrewed Aerial Vehicle}
\newacronym{fspl}{Fte.}{Friis Transmission Equation}
\newacronym{wifi}{Wi-Fi}{IEEE 802.11 Wireless Local Area Networks}
\newacronym{rssi}{RSSI}{Received Signal Strength Indicator}
\newacronym{mumimo}{MU-MIMO}{Multiple User Multiple Input Multiple Output}
\newacronym{ntp}{NTP}{Network Time Protocol}
\newacronym{uvdar}{UVDAR}{Ultraviolet Detection and Ranging}
\newacronym{lidar}{LiDAR}{Light Detection and Ranging}
\newacronym{ekf}{EKF}{Extended Kalman Filter}
\newacronym{imu}{IMU}{Inertial Measurement Unit}
\newacronym{fcu}{FCU}{Flight Control Unit}
\newacronym{rp}{RP}{Radiation Pattern}
\newacronym{aoa}{AoA}{Angle of Arrival}
\newacronym{aod}{AoD}{Angle of Departure}
\newacronym{aoi}{AoI}{Angle of Incidence}
\newacronym{uwb}{UWB}{Ultra Wideband}
\newacronym{itu}{ITU}{International Telecommunication Union}
\newacronym{eirp}{EIRP}{Equivalent Isotropically Radiated Power}

\newacronym{rmse}{RMS}{Root-Mean-Square-Error}
\newacronym{mae}{MAE}{Mean Absolute Error}
\newacronym{akaike}{AIC}{Akaike Information Criterion}
\newacronym{tlin}{LTD}{Linearized Time Demand}
\newacronym{quantile}{Q95}{95 Quantile}

\newacronym{methodMean}{M}{Mean}
\newacronym{methodNearestNeighbour}{NN}{Nearest Neighbor}
\newacronym{methodSphericalHarmonics}{SH}{Spherical Harmonics}
\newacronym{methodHarmonics}{AHS}{Analytical Harmonic Series}
\newacronym{methodPolynomial}{P}{Polynomial}
\newacronym{methodBucketized}{BG}{Basis Grid}

\newcommand{\addlinespace}[1][1]{}
\newcommand{\toprule}{\noalign{\hrule height 1.1pt}\noalign{\smallskip}}
\newcommand{\midrule}{\noalign{\smallskip}\noalign{\hrule}\noalign{\smallskip}\\[-1em]}
\newcommand{\bottomrule}{\noalign{\smallskip}\noalign{\hrule}\noalign{\smallskip}}

\DeclareSIUnit\decibelisotropic{dBi}

\begin{document}

\title{Full 3D Radiation Pattern Approximation in Multi-Rotor \acsp{uav}}
\hypersetup{pdftitle={Full 3D Radiation Pattern Approximation in Multi-Rotor UAVs}, pdfauthor={Martin Zoula, Daniel Bonilla Licea, Vaclav Navratil, Jan Faigl and Martin Saska}}

\author{Martin Zoula$^{1}$, Daniel Bonilla Licea$^{2}$, V\'{a}clav Navr\'{a}til$^{1}$, Jan Faigl$^{1}$, Martin Saska$^{1}$
\thanks{
  The work of M. Zoula was funded by the Czech Science Foundation (GAČR) under research project no. 25-17779M and CTU grant no. SGS26/077/OHK3/1T/13.
  The work of J. Faigl has been supported by MEYS of the Czech Republic under project No. LUABA24064.
  The work of M.~Saska was funded by the European Union under the project Robotics and advanced industrial production (reg. no. CZ.02.01.01/00/22\_008/0004590).}
\thanks{$^{1}$Authors are with the Faculty of Electrical Engineering, Czech Technical University in Prague, {\tt \{zoulamar|}{\tt navrava2|}{\tt faiglj|} {\tt saskam1\}@fel.cvut.cz}}
\thanks{$^{2}$D. B. Licea is with the Mohammed VI Polytechnic University, Morocco, {\tt daniel.bonilla@um6p.ma}}%
}%

  \thispagestyle{empty}
  \onecolumn
  {
  \topskip0pt
  \vspace*{\fill}
  \centering
  \LARGE{%
    This work has been submitted to the IEEE for possible publication. Copyright may be transferred without notice, after which this version may no longer be accessible.
    }
     \vspace*{\fill}
 }
 \NoBgThispage
 \twocolumn             
 \BgThispage

\markboth{IEEE Communications Letters}%
{Full 3D Radiation Pattern Approximation in Multi-Rotor \acsp{uav}}

\maketitle

\begin{abstract}
  In-flight outdoor approximation of full 3D \acp{rp} in non-holonomic multi-rotor \acp{uav} enables efficient communication-aware robotic missions.
  Our method identifies the \acl{methodSphericalHarmonics} series coefficients, decoupling the observed signal power gains into respective \acp{rp}.
  The proposed calibration trajectory provides training and testing samples of signal strength at an obstacle-free altitude.
  Simulation verifies the proposed novel approximation of the \acp{rp}, and real-world validation achieves an \acs{rmse} prediction error under 5\,dB.
\end{abstract}
\glsresetall

\section{Introduction} 

\IEEEPARstart{O}{ur} letter revises practical communication quality modeling in terms of \acp{rp} approximated from calibration flight with two or three \acp{uav}.
We explore a learnable \ac{rp} approximation model based on the Friis transmission equation, adequate for \acp{uav} in obstacle-free space:
\begin{equation}\label{eq:friis}
  \underbrace{\power_\rx = \power_\tx}_{\text{sig. power}} + \underbrace{\directivity_\rx + \directivity_\tx}_{\text{antenna gains \directivityjoint}} + \underbrace{20\log_{10}\frac{\wavelength}{4\pi\radiodistance}}_{\text{path loss}}.
\end{equation}
$\power$ are received and transmitted signal powers in \si{\deci\bel}, $\directivityjoint = \directivity_\tx + \directivity_\rx$ are (mis)aligned antenna \ac{rp} gains, $\radiodistance$ is the Euclidean distance between radios, and $\wavelength$ is wavelength.

\begin{figure}[!tb]
  \centering
  \includegraphics[width=\linewidth]{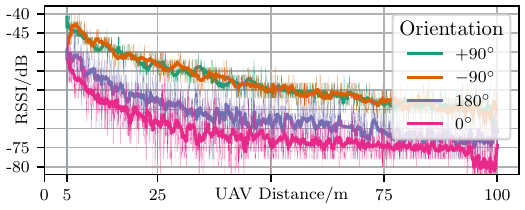}
  \includegraphics[width=\linewidth]{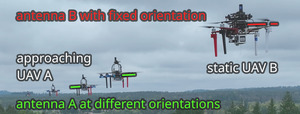}

  \caption{Illustration of our work. \ac{uav} A approaches \ac{uav} B four times; each time, however, attaining a different relative antenna orientation.
    Both antennas were always horizontal, parallel to the ground.
    However, antenna~A bore a different heading on each approach.
    As apparent from the recorded \acf{rssi}, choosing an appropriate orientation leads to consistent signal strength gain of around \SI{10}{\decibel}.
    }
  \label{fig:intro}
\end{figure}

While the \ac{rp} variation incurs up to \SI{-20}{\decibel} predictable attenuation~\cite{2017_rizwan_impact_of_uav_structure}, see also \cref{fig:intro}, its model can support robotic missions such as reconnection~\cite{2021_muralidharan_communication_aware_robotics__exploiting_motion_for}, connectivity-aware path planning~\cite{2019_mardani_communication_aware_uav_path_planning}, or transmission security~\cite{2024_bonilla_harnessing_the_potential_of_omnidirectional}.
However, field robots lack precise, up-to-date electromagnetic models~\cite{2025_diller_communication_mapping_for_robot}, thereby degrading rigorous approaches such as electromagnetic field solvers for autonomous onboard assessments.
Surrogate models or generic motion-communication frameworks are thus sought~\cite{2022_gielis_critical_review_of_communications_in} while
unpredictable radio neccesitates communication-less solutions in \ac{uav} swarming~\cite{2024_horyna_fast_swarming_of_uavs_in}.

Modular \ac{uav} systems enable autonomous multi-robot tasks with diverse sensors, actuators, and payloads~\cite{2023_hert_mrs_drone__a_modular_platform}.
However, ad~hoc reconfiguration changes the resulting \ac{rp} and necessitates re-identification to keep subsequent reasoning sound.
Similarly, as satellites degrade, their \acp{rp} change over time, necessitating re-identification~\cite{2024_mittermayer_novel_approach_for_in_orbit_satellite}.
Traditional antenna identification requires a calibrated probe antenna aligned towards the antenna under test in a controlled environment.
Neither is available in practical field deployments; moreover, both \ac{uav}-antenna systems need to identify themselves simultaneously.
Although contemporary aerial robotics already considers advanced electromagnetic phenomena~\cite{2019_yan_a_comprehensive_survey_on_uav}, \acp{rp} were so far considered only theoretically~\cite{2023_bonilla_licea_energy_efficient_fixed_wing_uav_relay_with} or with a fixed \ac{rp} model~\cite{2025_zhang_improving_data_collection_efficiency} to the best of our knowledge.

Our letter revises the \ac{uav} \ac{rp} identification~\cite{2021_matson_effect_of_antenna_orientation_on,2016_balanis_antenna_theory_analysis_and} and proposes a novel full 3D \ac{rp} approximation scheme.
\ac{rssi} samples of joint gain \directivityjoint{} are gathered along a synchronized calibration trajectory for two or three \acp{uav}.
Linear regression of \ac{methodSphericalHarmonics} coefficients decouples the observed joint $\directivityjoint$ into two independent \ac{rp} terms, $\directivity_\tx$ and $\directivity_\rx$.
We verify the approximation scheme computationally and in field experiments.
While two \acp{uav} learn the \ac{rp} up to the model's isotropic gain term, with three or more \acp{uav}, full 3D \acp{rp} are recovered.
Thus, we enable fast \ac{uav} \ac{rp} approximation outside the expensive and laborious anechoic chambers.

\section{Problem Statement}\label{sec:problem}

Let $\directivity(\azimuth,\inclination)$ be the unknown \ac{rp} of an antenna mounted on \ac{uav} A, while $\azimuth\in\ivalcc{0,2\pi},\inclination\in\ivalcc{-\frac{\pi}{2},\frac{\pi}{2}}$ are the azimuth and elevation of the \ac{aoi} \ac{wrt} \ac{uav}'s local reference frame.
In practice, the airframe and payload distort the radiated field~\cite{2020_badi_experimentally_analyzing_diverse_antenna_placements}, so each \ac{uav}-with-antenna assembly is treated as a single composite radiator~\cite{2024_hu_communications_channel_characteristics_in_the}.
Non-holonomic quadrotor \acp{uav} can maintain a heading $\yaw$ but not roll or pitch;
local $z$-axes are nominally vertical, reducing the freedom in desired pose control to $\vec{\statecontrol}(t)=\transpose{[x_\uav(t),y_\uav(t),z_\uav(t),\yaw_\uav(t)]}$~\cite{2024_bonilla_licea_when_robotics_meets_wireless_communications}.
However, onboard state estimation observes full state as $\vec{\state}(t) = \transpose{[x_\uav(\timephys),y_\uav(\timephys),z_\uav(\timephys),\roll_\uav(\timephys),\pitch_\uav(\timephys),\yaw_\uav(\timephys)]}$~\cite{2021_baca_the_mrs_uav_system__pushing}.
Bold symbols represent vectors and $\transpose{\cdot}$ is transposition.
Respective \ac{aoi} of UAV~B \ac{wrt} UAV~A $\azimuth^\uava_\uavb, \inclination^\uava_\uavb$ and vice versa are thus available as illustrated in \cref{fig:illustrate_directivity}.

\acp{uav} in an obstacle-free far-field link measure the joint directivity \directivityjoint{} as defined in \cref{eq:friis}.
The problem is to find (i) synchronized trajectories for \uavn{} \acp{uav} $\vec{\statecontrol}_\uava(t),\dots,\vec{\statecontrol}_\uavn(t)$ yielding informative training samples,
and (ii) a suitable parametric model for $\directivity$ capable of decoupling $\directivityjoint$ into $\directivity_\rx({\azimuth^\rx_\tx},{\inclination^\rx_\tx}) + \directivity_\tx({\azimuth^\tx_\rx},{\inclination^\tx_\rx})$,
while assuming matching polarization, a static and interference-free channel, and propeller-blade scattering absorbed into measurement noise.

\section{Proposed Method}\label{sec:method}\label{sec:method:trajectory}

\begin{figure}[!tb]
  \begin{tikzpicture}[x={(1cm,0cm)}, y={(0.4cm,0.4cm)}, z={(0cm,0.3cm)}]

  \def\ubx{3}   
  \def\uby{3}   
  \def\ubz{2}   

  \draw[dotted] (0,0,0) -- (\ubx,0,0) -- (\ubx,\uby,0);
  \draw[dotted] (0,0,\ubz) -- (0,\uby,\ubz) -- (\ubx,\uby,\ubz);
  \draw[dotted] (0,0,2) -- (\ubx,0,\ubz) -- (\ubx,\uby,\ubz);
  \draw[dotted] (\ubx,0,\ubz) -- (\ubx,0,0);
  \draw[dotted] (0,0,0) -- (0,\uby,0) -- (\ubx,\uby,0);
  \draw[dotted] (0,\uby,0) -- (0,\uby,\ubz);
  \draw[dashed] (\ubx,\uby,0) -- (\ubx,\uby,\ubz) -- (0,0,\ubz) -- (0,0,0);
  \draw[dashed] (0,0,0) -- (\ubx,\uby,\ubz) node[midway,fill=white] {\radiodistance};
  \draw[dashed] (0,0,0) -- (\ubx,\uby,0);

  \begin{scope}[shift={(0,0,0)}]


    \draw[line width=2pt,gray] ( 1, 0, 0) circle (.25);
    \draw[line width=2pt,gray] ( 0, 1, 0) circle (.25);
    \draw[line width=2pt,gray] (-1, 0, 0) circle (.25);
    \draw[line width=2pt,gray] ( 1, 0, 0) -- (-1,0,0);
    \draw[line width=2pt,gray] (0,-.4, 0) -- (0,1,0);

    \draw[red,  ->,line width=2.1pt] (0,0,0) -- (1.1,0,0);
    \draw[green!90!black,->,line width=2.1pt] (0,0,0) -- (0,1.1,0);
    \draw[blue, ->,line width=2.1pt] (0,0,0) -- (0,0,1.1);
    \draw (0,-1.1,0) node[] {UAV \uava, $\directivity_\uava$};

    \draw[thick] (0,0,0) -- (-.5,+.25,0) -- (-.5,+.25,.5);
    \draw[thick] (-.6,+.25,.4) -- (-.4,+.25,.4);
    \draw[thick] (-.6,+.25,.3) -- (-.4,+.25,.3);
    \draw[thick] (-.6,+.25,.2) -- (-.4,+.25,.2);

    \coordinate (b0) at (\ubx,\uby,\ubz);
    \coordinate (aa) at (\ubx,\uby,0);
    \coordinate (ai) at (\ubx,\uby,0);
    \coordinate (a0) at (0, 0, 0);
    \coordinate (ax) at (1, 0, 0);
    \pic [draw, ->, angle eccentricity=1.2, angle radius=1.6cm, "$\azimuth_\uavb^\uava$"] {angle = ax--a0--aa};
    \pic [draw, ->, angle eccentricity=.9, angle radius=3.5cm, "$\inclination_\uavb^\uava$"] {angle = aa--a0--b0};
  \end{scope}

  \begin{scope}[shift={(\ubx,\uby,\ubz)}, scale=0.8]


    \draw[line width=2pt,gray] ( 1, 0, 0) circle (.25);
    \draw[line width=2pt,gray] ( 0, 1, 0) circle (.25);
    \draw[line width=2pt,gray] (-1, 0, 0) circle (.25);
    \draw[line width=2pt,gray] ( 0,-1, 0) circle (.25);
    \draw[line width=2pt,gray] ( 1, 0, 0) -- (-1,0,0) (0,-1, 0) -- (0,1,0);
    \draw[red,  ->,line width=2.1pt] (0,0,0) -- (-1.1,0,0);
    \draw[green!90!black,->,line width=2.1pt] (0,0,0) -- (0,-1.1,0);
    \draw[blue, ->,line width=2.1pt] (0,0,0) -- (0,0,1.1);

    \draw (.5,0,5) node[] {UAV \uavb, $\directivity_\uavb$};

    \coordinate (a0) at (-\ubx,-\uby,-\ubz);
    \coordinate (ba) at (-\ubx,-\uby, 0);
    \coordinate (bi) at (-\ubx,-\uby, 0);
    \coordinate (b0) at (0, 0, 0);
    \coordinate (bx) at (-1, 0, 0);
    \pic [draw, ->, angle eccentricity=1.2, angle radius=1.6cm, "$\azimuth_\uava^\uavb$"] {angle = bx--b0--ba};
    \pic [draw, ->, angle eccentricity=.9, angle radius=3.5cm, "$\inclination_\uava^\uavb$"] {angle = ba--b0--a0};

    \draw[thick] (0,0,0) -- (.5,.25,0.) -- (.5,.25,.5);
    \draw[thick] (.4,.25,.2) -- (.6,.25,.2);
    \draw[thick] (.4,.25,.3) -- (.6,.25,.3);
    \draw[thick] (.4,.25,.4) -- (.6,.25,.4);

  \end{scope}

  \def\cx{1.5}   
  \def\cy{1.5}   
  \def\cz{1}   
  \def\r{2}  

  \draw[->, very thick, magenta, variable=\t, domain=10:-120, samples=50]
    plot ( {\cx + \r*cos(\t)}, {\cy + \r*sin(\t)}, {\cz+ \r*sin(\t)} );

  \draw[->, very thick, magenta, variable=\t, domain=190:60, samples=50]
    plot ( {\cx + \r*cos(\t)}, {\cy + \r*sin(\t)}, {\cz+ \r*sin(\t)} );

  \draw[very thin, magenta, variable=\t, domain=0:360, samples=100]
    plot ( {\cx + \r*cos(\t)}, {\cy + \r*sin(\t)}, {\cz+ \r*sin(\t)} );

  \def\cx{0}   
  \def\cy{0}   
  \def\cz{4}   
  \def\r{.5}  
  \draw[->, very thick, magenta, variable=\t, domain=0:300, samples=50]
    plot ( {\cx + \r*cos(\t)}, {\cy + \r*sin(\t)}, {\cz} );

  \draw[->, very thick, teal] (0,-.5,1) -- (-1.5,1,3.5) -- (0,2,4) -- (-3,5,3);
  \draw[->, very thick, teal] (3,3.5,1) -- (4.5,2,-1.5) -- (3,1,-1.5) -- (4.5,2,-2.5);
  \draw[->, very thick, teal] (-.5,4.5,3) -- (1.5,5,3) -- (.5,4.5,2.75) -- (1,4.5,4);

  \def\cx{3}   
  \def\cy{3}   
  \def\cz{4.5}   
  \def\r{.5}  
  \draw[->, very thick, magenta, variable=\t, domain=0:300, samples=50]
    plot ( {\cx + \r*cos(\t)}, {\cy + \r*sin(\t)}, {\cz} );

  \begin{scope}[shift={(-1,5,2)}, scale=0.4]

    \draw[line width=2pt,gray] ( 1, 0, 0) circle (.25);
    \draw[line width=2pt,gray] ( 0, 1, 0) circle (.25);
    \draw[line width=2pt,gray] (-1, 0, 0) circle (.25);
    \draw[line width=2pt,gray] ( 0,-1, 0) circle (.25);
    \draw[line width=2pt,gray] ( 1, 0, 0) -- (-1,0,0) (0,-1, 0) -- (0,1,0);
    \draw[red,->,line width=2.1pt] (0,0,0) -- (-1.1,0,0);
    \draw[blue, ->,line width=2.1pt] (0,0,0) -- (0,0,1.1);
    \draw[green!90!black,  ->,line width=2.1pt] (0,0,0) -- (0,-1.1,0);

    \draw (-1,1,1.5) node[] {UAV \uavc, $\directivity_\uavc$};

  \end{scope}

\end{tikzpicture}

  \caption{\acp{uav} see each other at azimuths $\azimuth_\uavb^\uava$ and $\azimuth_\uava^\uavb$.
  Although elevations $\inclination_\uavb^\uava$ and $\inclination_\uava^\uavb$ are opposite due to non-holonomy, we still consider both.
  Two \acp{uav} sample joint gain \directivityjoint{} along a common vertical circle of diameter \radiodistance{} (magenta). After each loop, one \ac{uav} changes its heading.
  Three \acp{uav} sample along randomized trajectories (cyan) with varying mutual distance.
  }
  \label{fig:illustrate_directivity}
  \label{fig:trajectory}
\end{figure}

While quadcopter \acp{uav} cannot maintain an arbitrary 6-\ac{dof} pose, we cannot directly emulate the anechoic chamber midair,
where one \ac{uav} would loiter in place, and the other would twist in place to cover all the directions.
We thus propose calibration trajectories to sample \directivityjoint{} over varying relative poses as illustrated in \cref{fig:trajectory}.

\subsubsection*{2-\acp{uav} Trajectory}
The $\left[\azimuth_\tx^\rx, \inclination_\tx^\rx, \azimuth_\rx^\tx, \inclination_\rx^\tx\right]\in\numReal^4$ space is impractical to sample fully.
That is why both \acp{uav} begin $\radiodistance=\SI{10}{\meter}$ away at an initial heading $\yaw_0$ and an initial height $z_0=\SI{20}{\meter}$, high enough to consider ground reflections negligible~\cite{2021_supramongkonset_empirical_path_loss_channel_characterization}.
Then, both \acp{uav} start flying around a vertical circle of a constant diameter~\radiodistance{}, centered directly between the \acp{uav} as illustrated in \cref{fig:trajectory}.
After each of $n$ loops, the first \ac{uav} changes its heading angle by $\Delta \yaw = \frac{\SI{360}{\degree}}{n}$.
After initial $n$ loops, the procedure repeats with the second \ac{uav} changing the headings.
Therefore, the \azimuth{}-\inclination{} is sampled in uniformly spaced meridians.
Trajectory units are used in a validation scheme, see \cref{sec:results:twouav}.

\subsubsection*{3-\acp{uav} Trajectory}
Generic randomized trajectories were used for 3-\acp{uav} calibration as illustrated in \cref{fig:trajectory}.
A sufficiently large bounding box in common Cartesian coordinates begins at a minimum altitude of \SI{15}{\meter}.
A trajectory of desired length or duration is constructed for the first \ac{uav} by randomly drawing positions in the bounding box and random headings.
A trajectory is constructed to pass through the specified waypoints under the desired dynamic constraints.
Next \ac{uav} does the same while rejecting the next random waypoint if the resulting trajectory temporarily approaches trajectories of the previous \ac{uav} too closely.
Although a randomized trajectory sacrifices strict predictability, its simplicity enables easy scalability to multiple \acp{uav} without compromising good sampling coverage across all \azimuth{}-\inclination{} spaces.

\subsubsection*{Approximation}
Observed samples of $\hat\directivityjoint_i$ are used in decoupling $\directivityjoint$ into components $\directivity_\rx + \directivity_\tx$.
($i$ is the index of a measurement.)
The \ac{rp} approximation is a residual minimization task over respective \ac{rp} parameters $\vec{\param}$ defined as
\begin{equation}
  \vec{\param}^\ast = \argmin_{\vec{\param}} \sum_i{(\hat\directivityjoint_i - \directivityjoint_i)^2}.
\end{equation}

\acl{rp} \directivity{} is a scalar function defined on a sphere over two coordinates $\azimuth$ and $\inclination$.
The scalar \acf{methodSphericalHarmonics} is a numerically advantageous series of regularized Legendre polynomials, decomposing spherical functions into basal components.
\ac{methodSphericalHarmonics} was already proven as a suitable tool in full 3D \ac{rp} modeling~\cite{2003_allard_the_model_based_parameter_estimation}:
\begin{multline}
  \directivity(\azimuth,\inclination;\vec{\param'}) = \sum_{l,m}\param_{l,m}Y_{l,m},\quad Y_{l,m} =\\
  \begin{cases}
    -1^m \sqrt{2\frac{2l+1\fact{(l-\abs{m})}}{4\pi\fact{(l+\abs{m})}}} \legendre^\abs{m}_l(\cos\azimuth)\sin(\abs{m}\inclination), & m < 0 \\
    \sqrt{\frac{2l+1\fact{(l-\abs{m})}}{4\pi\fact{(l+\abs{m})}}} \legendre^m_l(\cos\azimuth), & m = 0 \\
    -1^m \sqrt{2\frac{2l+1\fact{(l-m)}}{4\pi\fact{(l+m)}}} \legendre^m_l(\cos\azimuth)\sin(\abs{m}\inclination), & m > 0,
  \end{cases}
\end{multline}
where $\legendre$ is the associated Legendre polynomial.
Parameter values $\param_{l,m}\in\vec{\param'}\in\vec{\param}$ are thus found as solutions to an ordinary overdetermined linear least squares problem.
The number of parameters relates to the \ac{methodSphericalHarmonics} series truncation; series capped by $l\leq N$ is further denoted as \ac{methodSphericalHarmonics}-N.

\section{Simulation}\label{sec:results}
\label{sec:decoupling}

We assessed the proposed antenna decoupling scheme for parameter identifiability in simulation.
Given a sample of
\begin{equation}
  \power_{\rx,i}-\power_{\tx,i}-20\log_{10}\frac{\wavelength}{4\pi\radiodistance_i} = \directivityjoint_i \sim \left[\azimuth_\uavb^\uava,\inclination_\uavb^\uava,\azimuth_\uava^\uavb,\inclination_\uava^\uavb\right]_i,
\end{equation}
\ac{methodSphericalHarmonics} transforms $(\azimuth$-$\inclination)_i$ pairs to resp. feature vectors $\vec{\features}_i$.
$\vec{\features}(\azimuth,\inclination) = \left[Y_{0,0}(\azimuth,\inclination), Y_{1,-1}(\azimuth,\inclination), \dots\right]$ with angles matching the transmission counterpart.
Parameters $\vec{\param}$ are expressed as per-\ac{uav} vector concatenation $\transpose{\left[\vec{\parama}, \vec{\paramb}, \vec{\paramc}, \dots\right]}$, assuming an arbitrary \ac{uav} count.
The equation is arranged as
\begin{equation}
  \begin{bmatrix}
    \vec\features^\uava_1, &\vec\features^\uavb_1, &\vec0,            &\dots\\
    \vec\features^\uava_2, &\vec0,                 &\vec\features^c_2 &\dots\\
    \vdots             & \vdots            &\vdots        &\ddots
  \end{bmatrix}
  \begin{bmatrix}
    \vec\parama~
    \vec\paramb~
    \vec\paramc~
    \dots
  \end{bmatrix}^\transposesym{}
  =
  \begin{bmatrix}
    \hat{\directivityjoint}_1\\
    \hat{\directivityjoint}_2\\
    \vdots
  \end{bmatrix}.
\end{equation}
Each row of the left matrix corresponds to a signal sample and contains only features of the two antennas involved.

The parameter identifiability was analyzed with surrogate antenna radiation patterns according to the \ac{itu} Recommendation \texttt{ITU-R F.1336-5}~\cite{2019_itu_r_f_1336_5}.
The signal samples were drawn as pairwise transmissions at various azimuths and elevations.

\begin{figure}[!ht]\centering

  \subfloat[
    Constant prediction error with $\param_{0,0}\pm\epsilon$.
    Error raises with $\epsilon > 10^5$ due to insufficient numerical precision.
    \label{fig:decoupling_illus_2uav:errors}
  ]{\includegraphics[width=\linewidth]{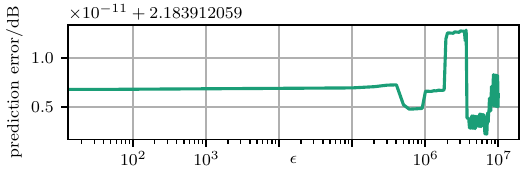}}

  \subfloat[
    The \ac{itu} circular (``C'') and sectoral (``S'') were selected as ground truth (``GT''). Adding $\pm\epsilon$ to $\param_{0,0}$ in respective learned \ac{methodSphericalHarmonics}-8 models does not change the prediction error.
    \label{fig:decoupling_illus_2uav:patterns}
  ]{\includegraphics[width=\linewidth]{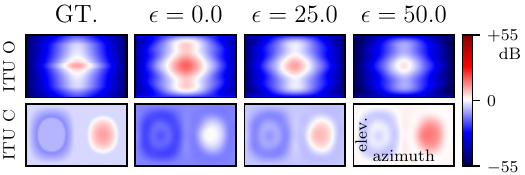}}

  \caption{Rank-deficient two-antenna approximation.
  }
  \label{fig:decoupling_illus_2uav}
\end{figure}

If two \acp{uav} are considered, the rank of the feature matrix is one less than the number of columns.
The null space of the matrix exactly identifies the parameters related to $Y_{0,0}$ of both antennas, i.e., the modeled isotropic gain.
Hence, adding an unitless $\epsilon$ to $\parama_{0,0}$ and $-\epsilon$ to $\paramb_{0,0}$ does not change the prediction error as depicted in \cref{fig:decoupling_illus_2uav}.
($\parama_{0,0}$ and $\paramb_{0,0}$ are multiplicative parameters of $Y_{0,0}$ for respective \acp{uav}.)
Ridge regression heuristically minimizes the parameter weights to obtain a unique solution.
Thus, shapes of both \acp{rp} are recovered, but $Y_{0,0}$, the model's isotropic gain (bias), remains unknown.
We argue that this approximation already enables important robotic applications such as orienting the \acp{uav} toward more favorable directions.

\begin{figure}[!ht]\centering
  \includegraphics[width=\linewidth]{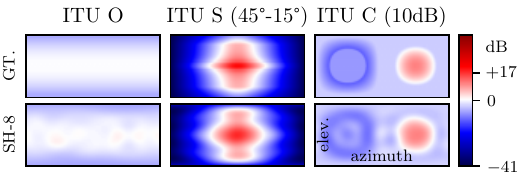}
  \caption{Illustration of the full-rank decoupling with three ITU antenna models.
    Three ITU antenna models (denoted as Ground Truth or ``GT'') were used, the default omnidirectional (``O''), sectoral with \SI{45}{\degree} and \SI{15}{\degree} \SI{3}{\decibel} zone and \SI{20}{\decibel} gain (``S'') and circular with \SI{10}{\decibel} gain (``C'').
  }
  \label{fig:decoupling_illus}
\end{figure}

The rank deficiency is resolved by adding at least a~third antenna, as mentioned in the fundamental textbook~\cite{2016_balanis_antenna_theory_analysis_and}.
We depict an example simulation of three antenna models in \cref{fig:decoupling_illus}.
However, both methods strictly assume co-polarized radiation as the relative ``roll'' angle is not taken into account.

\section{Real World Experiments}\label{sec:results:setup}

The models were evaluated on datasets captured during several real-world outdoor flights, using off-the-shelf \SI{2.4}{\giga\hertz} radios.
The prediction \acs{rmse} error was under \SI{5}{\decibel}.
The \acl{methodSphericalHarmonics} demonstrated cross-experiment generalization with the best-performing model order.

\begin{figure}[!b]\centering
  \newcommand{\localimageheight}{2.1cm}
  \subfloat[
    Footage from a calibration flight execution. \acp{uav} circle around each other, and after each loop, one \ac{uav} slightly changes its heading.
    \label{fig:dataset_illus}
  ]{\includegraphics[height=\localimageheight]{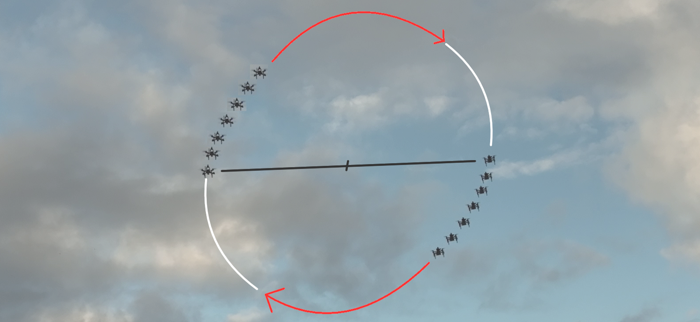}}
  \hfill
  \subfloat[
    Two \texttt{Holybro X-500} \ac{uav} with \texttt{T2U} antennas (marked with red ellipses) in vertical position. Miscellaneous sensors attached.
    \label{fig:experiment_illustration:x500}
  ]{\includegraphics[height=\localimageheight]{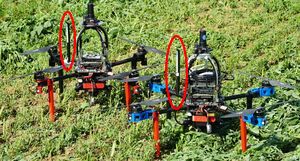}}

  \caption{Used \ac{uav} platforms and the experimental venue.}
  \label{fig:experiment_illustration}
\end{figure}

The trajectories were executed by quadrotor \acp{uav} built around the \texttt{Holybro X500 V2} frame and \texttt{Holybro Pixhawk 6X} \ac{fcu} with miscellaneous payloads.
See \cref{fig:experiment_illustration}.
The MRS UAV System~\cite{2021_baca_the_mrs_uav_system__pushing} utilized the state estimates by a 24-\ac{dof} \ac{ekf} via \texttt{Holybro H-RTK F9P GNSS} module, magnetometer, \acs{imu}, and barometer~\cite{2015_meier_px4__a_node_based_multithreaded_open}.
\acp{uav} were powered by a 4-cell serial lithium-polymer battery providing around \SI{10}{\min} of mission time.

\label{sec:results:radio}
The \ac{rssi} was recorded by off-the-shelf \texttt{TP Link Archer T2U Plus} \acs{wifi} USB modules with \SI{5}{\decibelisotropic} monopole antenna.
The \ac{rssi} is available via so-called ``monitor mode.''
Antennas were oriented vertically \ac{wrt} the \ac{uav} frame to decouple the experiment from polarization losses, as shown in \cref{fig:experiment_illustration:x500}.
Devices exchanged frequent probe messages, each implicitly annotated with the nominal \SI{20}{\decibel\milli\null} \ac{eirp}.

\subsection{Two \acp{uav}}
\label{sec:results:twouav}

The two-\acp{uav} dataset draws from a flight over an open, slightly sloped field in a rural area.
The nearest significant obstacle (a solitary tree) was about \SI{40}{\meter} away.
The weather was windy, with an air speed of around \SI{10}{\meter\per\second};
the disturbances are reflected in the onboard state estimation.

\begin{figure}[ht!]

  \subfloat[
    Mutual relative bearing of both \acp{uav} as azimuth-elevation plots, captured during the experimental campaign.
    Green and orange represent independent training and testing samples, respectively.
    (Signal strength not depicted.)
  ]{\includegraphics[width=\linewidth]{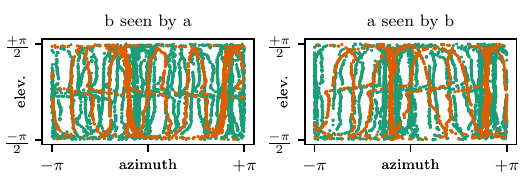}}

  \vspace{1em}
  \subfloat[
    Trajectory and $\power_\rx$ over time in one \ac{uav}.
    $x$ axis complements the circle, while $y$ is constant zero.
    $\power_\rx$ plateau at the beginning is an insignificant measurement artifact.
  ]{\includegraphics[width=\linewidth]{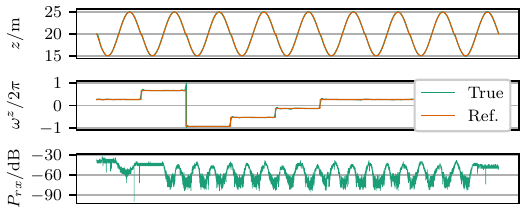}}
  \vspace{1em}

  \caption{
    Illustration of the 2-\acp{uav} dataset.
  }
  \label{fig:dataset} \label{fig:dataset_cover}
\end{figure}

The proposed methods were validated in a learning-testing scheme.
\Acp{uav} gathered samples along particular trajectory units from \cref{sec:method:trajectory}:
By setting the loop count $n=5$, the $\Delta\yaw$ becomes $\SI{72}{\degree}$.
Three learning trajectory units with $\yaw_0 \in \left[\SI{0}{\degree}, \SI{24}{\degree}, \SI{48}{\degree}\right]$ were concatenated, providing \num{56164} learning \ac{rssi} samples.
One trajectory unit with $\Delta\yaw = \SI{36}{\degree}$ yielded \num{24087} test data samples separated from the training samples in the 4D relative configuration space.
The resulting coverage of $\azimuth$-$\inclination$ spaces for both \acp{uav} is depicted in \cref{fig:dataset}.

Testing samples were evaluated for prediction performance using \ac{rmse} and \ac{akaike}.
Further, a rough estimate of computational demand is provided as \ac{tlin}, the total CPU time required for learning, measured by Python's \texttt{time.perf\_counter}.
Finally, we include the absolute errors \ac{quantile}.

\begin{figure}[!ht]
  \centering
  \vspace{1em}
  \includegraphics[width=\linewidth]{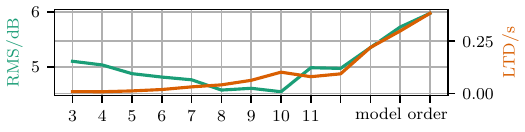}
  \caption{Model order selection: Green curve depicts prediction error (\ac{rmse}), identifying the \ac{methodSphericalHarmonics}-10 model as the most potent.
   The orange curve illustrates increasing time demands as model order increases.}
  \label{fig:model_order_selection}
  \vspace{1em}
\end{figure}

\begin{figure}[!ht]\centering
  \newcommand{\lwpara}{.5}

  \null
  \hfill
  \subfloat[\label{fig:result:sh10}\acs{methodSphericalHarmonics}-10, optimal order.]{\includegraphics[width=\lwpara\linewidth]{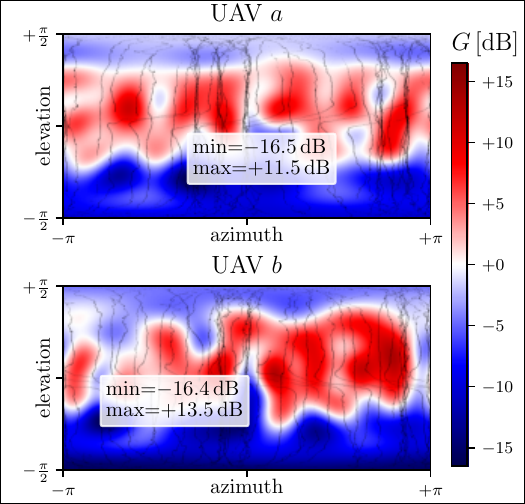}}
  \hfill
  \subfloat[\label{fig:result:sh15}\acs{methodSphericalHarmonics}-15, sampling bias.]{\includegraphics[width=\lwpara\linewidth]{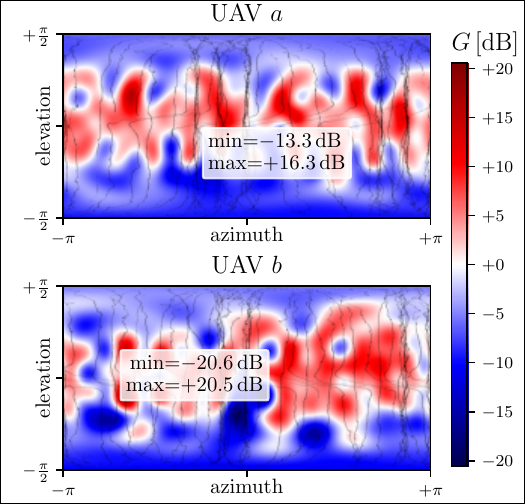}}
  \hfill
  \null

  \caption{A learned radiation patterns selection. The sub-figures pairs depict both decoupled \acp{rp} $\directivity_\uava$ and $\directivity_\uavb$ in the \azimuth{}-\inclination{} domain.
    Black points represent the samples, same as in \cref{fig:dataset_cover}. The denser sampling at certain UAV~A's azimuths corresponds to the sparse sampling on UAV~B and vice versa.
  }
  \vspace{1em}
  \label{fig:result_patterns}
\end{figure}

The independent test trajectory unit enabled selecting the best model order based on test error.
\Cref{fig:model_order_selection} shows the prediction error for $l\in\left[3, \dots, 15\right]$.
The \num{8}\textsuperscript{th} to \num{10}\textsuperscript{th} orders exhibit the lowest prediction error and are likely the most suitable for learning the \acp{rp}.
Finally, \cref{fig:result:sh15} depicts a higher \num{15}\textsuperscript{th} order \ac{methodSphericalHarmonics}, which is already overfitted, as also indicated by the increased \ac{akaike}.
We further consider only the \ac{methodSphericalHarmonics}-10.
Resulting radiation patterns are illustrated in \cref{fig:result_patterns}.

The selected baseline models are the feature-agnostic \ac{methodMean} and the \acf{methodNearestNeighbour} regressions.
Besides, basic \ac{methodPolynomial} regression disregards the periodicity of the spherical domain, and \ac{methodBucketized} filters representative samples for each \ac{rp}.
The baseline \acf{methodMean} method provided an upper bound on prediction accuracy of \SI{11.15}{\decibel} \ac{rmse}.
Contrarily, the \num{400}-neighbor \ac{methodNearestNeighbour} provided a hint on the best possible prediction of \SI{4.82}{\decibel} \ac{rmse}.

The \ac{methodSphericalHarmonics}-10 method emerged as the most promising:
It attained the lowest prediction error without requiring any hyperparameter tuning, unlike the \ac{methodBucketized} baseline.
Finally, the \ac{methodPolynomial} method proves to be less useful, as it ignores the domain periodicity.
Besides, the condition number was \num{122.58}, indicating the problem was well-conditioned and that the training trajectory truly covered both $\azimuth$-$\inclination$ spaces and decoupled the \acp{rp}.
The learned \acp{rp} correspond to the omnidirectional pattern as expected from the antenna placement.

\begin{table}[!ht]\centering
  \caption{Performance comparison of selected methods.}
  \label{tbl:model_comparison}
  \begin{tabular*}{\linewidth}{@{\extracolsep{\fill}}lrrrrr}
\toprule
  Method & \ac{rmse}{\scriptsize $/\si{\decibel}$} & $\abs{\parama}^\dagger$ & \ac{akaike}{\scriptsize $/10^3$}& \ac{tlin}{\scriptsize $/\si{\second}$}& Q95{\scriptsize $/\si{\decibel}$}\\
\midrule\addlinespace[2.5pt]\\[-1em]
M$^\ast$ & 11.15 & 1 & 116.16 & 0.0003 & 18.22 \\
NN-400$^\ast$ & 4.82 & 56,164$^\ddagger$ & 123.96 & 0.0235 & 9.35 \\
SH-10 & 4.56 & 200 & 73.05 & 0.0808 & 8.52 \\
P-12 & 4.91 & 182 & 76.69 & 0.0583 & 9.21 \\
BG-7 & 4.81 & 196 & 75.64 & 0.0699 & 8.94 \\
\bottomrule
\multicolumn{6}{l}{\scriptsize$\null^\dagger$ Number of parameters per a model of a single antenna.}\\
\multicolumn{6}{l}{\scriptsize$\null^\ddagger$ Number of \ac{methodNearestNeighbour} parameters equals the number of samples in the training dataset.}\\
\multicolumn{6}{l}{\scriptsize$\null^\ast$ Baseline methods incapable of \ac{rp} decoupling.}
\end{tabular*}

\end{table}

\subsection{Three \acp{uav}}

\begin{figure}[!ht]
  \centering
  \subfloat[
    Model order selection: The green curve depicts test prediction error (\ac{rmse}), indicating that the \ac{methodSphericalHarmonics}-3 model is the most effective when using all three \acp{uav}.
    The orange curve illustrates the increasing condition number, while the purple curve shows the unitless \ac{akaike}.
  ]{\includegraphics[width=\linewidth]{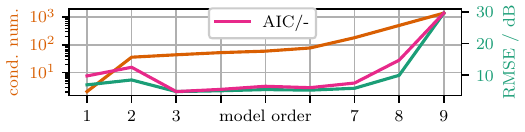}}

  \subfloat[
    Azimuth-elevation coverage for the randomized training trajectory.
  ]{\includegraphics[width=\linewidth]{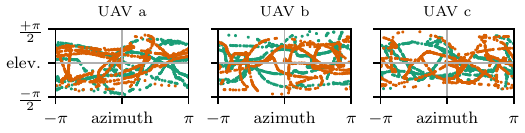}}

  \subfloat[
    Learned \acp{rp}, exactly corresponding to the intuition of omnidirectional radiator with variations because of \ac{uav} structure.
    \label{fig:model_order_selection_3uav:aeplot}
  ]{\includegraphics[width=\linewidth]{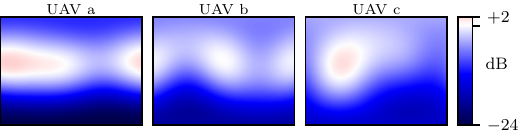}}

  \subfloat[
    The same as above, plotted in 3D. Radius corresponds to the power quotient, not \si{\decibel}. Colors and labels match \cref{fig:model_order_selection_3uav:aeplot}.
  ]{\includegraphics[width=\linewidth]{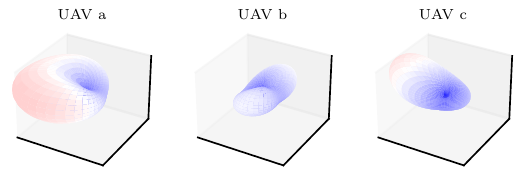}}

  \caption{Results of real world experiment when 3 \acp{uav} learned their \acp{rp} from samples gathered along the randomized trajectory.}
  \label{fig:model_order_selection_3uav}
\end{figure}

The three-\acp{uav} experiment comprised two independent random trajectories, each lasting around \SI{5}{\minute}.
The trajectories were executed in an open field with the nearest obstacles more than \SI{50}{\meter} away.
However, the \ac{uav} payloads were different from those in the two-\ac{uav} setting.
The first flight served as a learning dataset with \num{21778} samples; the second as a test dataset with \num{11831} samples.
Already with \ac{methodSphericalHarmonics} as low as 3, we found an optimal model order with \ac{rmse} of \SI{4.8}{\decibel} and \ac{akaike} \num{34782}.
\Cref{fig:model_order_selection_3uav} summarizes the 3-\ac{uav} experiment.

The resulting \acp{rp} correspond to the intuition behind an omnidirectional radiator with the biggest gain along the equator, yet distorted by the \ac{uav} structure.
Recall that the 2-\ac{uav} case had a significantly higher optimal model order; we interpret this as a bias due to systematic oversampling of selected regions.
On the contrary, randomized sampling in a 3-\ac{uav} scenario overcame this limitation while remaining fundamentally well-determined.
In both cases, the \ac{rmse} came near to the inherent noise level in \ac{rssi} measurement of approx. \SIrange{2}{4}{\decibel}, depending on the transmitter.

\section{Conclusion}\label{sec:conclusion}

The results demonstrated the practical feasibility of simultaneously identifying full 3D \acp{rp} of co-polarized monopole antennas distorted by the \acp{uav}' structure.
After executing a specific synchronized calibration trajectory, the \acl{methodSphericalHarmonics} model yielded the least-test-error \ac{rp} parameter values.
This approach can be a building block for full-polarimetric identification, precision improvement, radio-based sensing~\cite{2024_mittermayer_novel_approach_for_in_orbit_satellite}, localization~\cite{2024_ibrahim_inferring_direction_and_orientation_from}, and communication-aware planning in autonomous robotics~\cite{2021_muralidharan_communication_aware_robotics__exploiting_motion_for}.

\balance

\printbibliography
\end{document}